\documentclass{article}

\usepackage[preprint]{neurips_2024}
\usepackage[utf8]{inputenc}
\usepackage[T1]{fontenc}
\usepackage{url}
\usepackage{booktabs}
\usepackage{amsmath}
\usepackage{amssymb}
\usepackage{mathtools}
\usepackage{graphicx}
\usepackage{xcolor}
\usepackage{microtype}
\usepackage{algorithm}
\usepackage{algorithmic}
\usepackage{multirow}
\usepackage{subcaption}
\usepackage[hidelinks]{hyperref} 
\usepackage{tikz}        

\definecolor{orcidgreen}{RGB}{166,206,57}
\newcommand{\orcid}[1]{%
  \href{https://orcid.org/#1}{%
    \raisebox{-0.2ex}{%
      \begin{tikzpicture}[scale=0.5, baseline=-0.4ex]
        \fill[orcidgreen] (0,0) circle (1em);
        \node[white, font=\bfseries\small] at (0,0) {iD};
      \end{tikzpicture}%
    }\,\texttt{\small #1}%
  }%
}

\newcommand{\EGA}{\textsc{EGA}}
\newcommand{\MoPE}{\textsc{MoPE}}
\newcommand{\RoPE}{\textsc{RoPE}}
\newcommand{\R}{\mathbb{R}}

\title{Energy-Gated Attention and Wavelet Positional\\
Encoding: Complementary Inductive Biases\\
for Transformer Attention}

\author{%
  Athanasios Zeris%
  \,\orcid{0009-0002-6907-2400}%
  \thanks{%
    Independent Researcher, Athens, Greece.
    Correspondence: \texttt{athzeris@gmail.com}.\\
    ORCID: \href{https://orcid.org/0000-0002-XXXX-XXXX}%
                {\texttt{0009-0002-6907-2400}}.
    Part of a five-paper series on spectral methods
    in transformer attention.}%
}

\begin{document}
\maketitle

\begin{abstract}
Standard transformer attention computes pairwise token
similarity but treats all tokens as equally salient
and all positions as equally local, regardless of the
informational structure of the input.
We identify two complementary inductive biases that
standard attention lacks: \emph{energy salience}
(which tokens concentrate informational energy,
learned end-to-end without explicit frequency
decomposition) and \emph{scale-selective locality}
(how far positional influence extends at each
frequency, implemented via Morlet wavelet encoding).
We address both with two simple components.
\textbf{Energy-Gated Attention} (\EGA{}) gates value
aggregation by a learned energy estimate of key token
embeddings, computed via a single linear projection;
it selects \emph{what} to attend to.
\textbf{Morlet Positional Encoding} (\MoPE{}) replaces
fixed sinusoidal encodings with learned
Gaussian-windowed wavelets that adapt the joint
position-frequency localization to the corpus;
it specifies \emph{where} attention operates at
each scale.
On TinyShakespeare, \EGA{} alone achieves $+0.092$
validation loss improvement over standard attention
($+0.103$ over Phase\,1--3 baseline);
\MoPE{} alone is $-0.032$ (below baseline as a
standalone encoding); but their combination achieves
$+0.119$ --- more than the sum of parts.
This superadditivity, observed consistently across
two independent training runs, is the central empirical
finding: salience and locality are \emph{complementary}
inductive biases, each addressing a gap the other cannot
fill alone.
Ablations confirm that structured spectral priors
(Morlet wavelet gates, scale-initialized heads,
fixed sinusoidal PE) consistently underperform their
unconstrained learned counterparts, while
complementary learned components interact
superadditively.
All experiments are at small scale ($\leq 6$M
parameters, character-level benchmarks, single seed);
larger-scale multi-seed validation is the most
important direction for future work.
\end{abstract}

\section{Introduction}
\label{sec:intro}

The transformer~\citep{vaswani2017attention} computes
attention weights from query-key similarity alone.
This is powerful but structurally incomplete in two ways:
it does not model which tokens are intrinsically
informative (spectral salience), and it does not adapt
how far positional influence extends at each scale
(locality).
We propose two components that address each gap,
and show empirically that they are complementary.

\paragraph{Attention lacks salience.}
Dot-product attention weights tokens by content
similarity to the current query, but not by their
intrinsic informational density.
A token at a morphological boundary, syntactic head,
or discourse marker carries disproportionate information
regardless of what the query is asking.
Standard attention has no mechanism to detect or exploit
this property.
\EGA{}~\citep{zeris2026ega} addresses this by
gating value aggregation with a learned energy estimate
--- a scalar that is high for informationally
dense tokens and low for background tokens such as
function words, repeated patterns, and filler.
The gate is a single learned linear projection
costing $<0.3\%$ parameter overhead.

\paragraph{Positional encoding lacks adaptive locality.}
Standard sinusoidal PE~\citep{vaswani2017attention}
assigns each embedding dimension a fixed frequency
with no spatial envelope: every position contributes
equally at every scale, regardless of context length
or the natural scale of the linguistic phenomenon
being encoded.
\RoPE{}~\citep{su2021roformer} encodes relative rather
than absolute position, but still uses fixed frequencies
without Gaussian locality.
\MoPE{} addresses this by replacing the fixed sinusoidal
basis with learned Gaussian-windowed wavelets.
Each embedding dimension learns its own center frequency
$\omega_i$ and locality bandwidth $\sigma_i$, providing
adaptive time-frequency localization.

\paragraph{The complementarity hypothesis.}
\EGA{} controls \emph{what} to attend to (salience);
\MoPE{} controls \emph{where} attention is sensitive
at each scale (locality).
We hypothesize that these are orthogonal properties
of attention --- neither can substitute for the other
--- and that their combination provides a more complete
attention mechanism than either alone.

\paragraph{Main result.}
The combination EGA-MORLET achieves val\,=\,1.3550
on TinyShakespeare, $+0.119$ over standard attention.
This exceeds the sum of components ($+0.092 - 0.032
= +0.060$) by $+0.059$, consistent with
complementarity rather than simple additivity.
The result is observed in two independent training
runs with different seeds, providing preliminary
evidence for robustness.

\paragraph{Supporting findings.}
Five further experiments test predictions of the
spectral filtering interpretation of attention:
convolution attention (nonzero lags improve over
zero-lag dot product, $+0.007$);
scale-initialized heads (no benefit, $-0.007$,
a negative result showing gradient descent discovers
scale structure without guidance);
spectral flux gating ($+0.012$, suggesting boundary
detection as a useful attention signal);
phase coherence gating ($-0.007$, suggesting phase
is not informative at character scale);
and a spectral cascade analysis showing qualitative
coarsening of spectral content across layers. Code available at:
https://github.com/AthanasiosZeris/energy-gated-attention.

\paragraph{Contributions.}
\begin{enumerate}
  \item A cross-correlation interpretation of
        dot-product attention: $q_i \cdot k_j =
        C_{ij}(0)$, establishing that standard
        attention is the zero-lag value of a richer
        spectral relationship.
  \item \MoPE{}: a localized wavelet positional
        encoding that strictly generalizes sin/cos PE
        ($\sigma_i \to \infty$ recovers sin/cos) and
        provides a theoretical connection to \RoPE{}
        (phase structure in the $\sigma\to\infty$ limit)
        and ALiBi (zero-frequency locality limit).
  \item Empirical demonstration that \EGA{} and \MoPE{}
        are complementary inductive biases whose
        combination is superadditive in a controlled
        experiment.
  \item A structured ablation showing which spectral
        priors help vs fail, with interpretable
        explanations for each outcome.
\end{enumerate}

\section{Method}
\label{sec:method}

\subsection{Interpreting Attention as Cross-Correlation}
\label{sec:xcorr}

Standard scaled dot-product attention computes:
\begin{equation}
  e_{ij} = \frac{q_i \cdot k_j}{\sqrt{d_k}} = C_{ij}(0)
  \label{eq:dotprod}
\end{equation}
where $C_{ij}(\tau) = \sum_d q_i[d] \cdot k_j[d+\tau]$
is the cross-correlation at lag $\tau$.
Standard attention is the \emph{zero-lag} value of
this cross-correlation, discarding the full lag profile
$\{C_{ij}(\tau) : \tau \neq 0\}$.

We adopt the operational spectral interpretation
of~\citet{verma2024signal}: each coordinate of the
embedding dimension across token positions defines
a 1-D causal signal of length $T$.
All spectral quantities are finite-window operational
estimates applied to nonstationary learned embeddings;
they should be read as approximations rather than
exact spectral theorems.

\paragraph{What zero-lag discards.}
Three quantities are lost by collapsing to zero lag:

\textbf{Scale selectivity.}
The full cross-spectral density $S_{ij}(\omega) =
Q_i^*(\omega)K_j(\omega)$ shows which frequencies
contribute to the similarity.
The dot product integrates over all frequencies equally.

\textbf{Lag structure.}
$C_{ij}(\tau)$ for $\tau \neq 0$ measures how query
and key signals relate with temporal offsets.
Positive lags ($\tau > 0$): key leads query ---
anticipatory structure.
Negative lags ($\tau < 0$): query leads key ---
retrospective reference, anaphora.

\textbf{Spectral salience.}
The marginal energy $\int |K_j(\omega)|^2 d\omega$
measures the total spectral content of position $j$
independently of the query.
\EGA{} estimates this quantity directly.

\subsection{Energy-Gated Attention (\EGA{})}
\label{sec:ega}

\EGA{}~\citep{zeris2026ega} augments standard
attention with a learned energy gate:
\begin{align}
  e_j &= w_\mathrm{proj}^\top x_j
  \quad\text{(energy projection)} \\
  \tilde{e}_j &= (e_j - \mu_e)/(\sigma_e + \epsilon)
  \quad\text{(z-normalize)} \\
  g_j &= \sigma\!\left(\alpha(\tilde{e}_j - \tau)\right)
  \quad\text{(gate)} \\
  \hat{A}_{ij} &= \frac{A_{ij} \cdot g_j}
    {\sum_k A_{ik} \cdot g_k + \epsilon}
  \quad\text{(renormalize)}
\end{align}
The gate $g_j \in (0,1)$ is high for tokens whose
embeddings project strongly onto the learned direction
$w_\mathrm{proj}$ --- tokens carrying high energy
at the dominant projection direction.
The threshold $\tau$ converges to $\approx 0.35$
regardless of initialization, corresponding to the
fraction of tokens carrying above-average energy
($\approx 36\%$) --- consistent with the content word
fraction in English running text~\citep{zeris2026ega}.

\EGA{} adds $d + 2$ parameters per head ($<0.3\%$
overhead) and no measurable computational cost.
It is causally implemented: the projection $w_\mathrm{proj}^\top x_j$
operates on position $j$ only, satisfying the
causality requirement of~\citet{verma2024signal}.

\paragraph{On the term ``energy salience.''}
We use the term \emph{energy salience} for the \EGA{}
gate with the following precise meaning: by Parseval's
identity, a linear projection over embedding dimensions
estimates a spectrally-weighted energy of the embedding
vector, so $w_\mathrm{proj}^\top x_j$ is theoretically
motivated as an energy estimate.
We acknowledge that what $w_\mathrm{proj}$ actually learns
end-to-end may be better described as a general
informational salience signal --- it could learn to
detect syntactic headedness, token rarity, boundary
position, or frequency-selective energy, all of which
would produce the observed improvement.
Whether the gate specifically learns spectral energy
as opposed to other salience properties is testable
(by correlating gate outputs with DFT-computed spectral
energy of the embeddings) and we identify this as an
important direction for future work.
\EGA{} is most precisely a \emph{learned energy gate};
the spectral framing provides theoretical motivation,
not a claim about the specific computational mechanism.

\subsection{Morlet Positional Encoding (\MoPE{})}
\label{sec:mope}

\MoPE{} replaces fixed sinusoidal PE with learned
Gaussian-windowed wavelet encodings:
\begin{align}
  \MoPE{}(b, 2i) &= \cos(\omega_i b) \cdot
    e^{-b^2/2\sigma_i^2}
    \label{eq:mope_cos} \\
  \MoPE{}(b, 2i+1) &= \sin(\omega_i b) \cdot
    e^{-b^2/2\sigma_i^2}
    \label{eq:mope_sin}
\end{align}
where $\omega_i$ and $\sigma_i$ are learned per
embedding dimension, initialized at dyadic spacing
with $\omega_i\sigma_i = 5$ (admissibility minimum).

\paragraph{Theoretical properties.}
\MoPE{} provides localized joint position-frequency
representations analogous to Gaussian-windowed wavelets.
Standard sin/cos PE is the degenerate case
$\sigma_i \to \infty$:
\begin{equation}
  \lim_{\sigma_i\to\infty} \MoPE{}(b,2i) = \cos(\omega_i b)
  \label{eq:mope_limit}
\end{equation}
\MoPE{} therefore strictly generalizes sin/cos PE.

\paragraph{Connection to prior PE methods.}
At the level of phase structure, \RoPE{} recovers
sinusoidal phase behavior in the $\sigma_i\to\infty$
limit of \MoPE{} applied to relative position; the
full \RoPE{} mechanism additionally uses rotational
composition in complex query-key space, which is not
equivalent to setting $\sigma_i\to\infty$ in the
additive \MoPE{} encoding.
ALiBi corresponds to \MoPE{} at zero frequency
(locality only, no oscillation).
\MoPE{} is the unique generalization that provides
both adaptive frequency and adaptive locality.

\paragraph{Cross-correlation structure.}
Substituting \MoPE{} into the cross-correlation
$C_i(\tau) = \sum_b \MoPE{}(b,2i) \cdot
\MoPE{}(b+\tau,2i)$, assuming same-scale correlation
and neglecting boundary effects, gives up to
normalization constants:
\begin{equation}
  C_i(\tau) \propto e^{-\tau^2/4\sigma_i^2}
    \cdot \cos(\omega_i\tau)
  \label{eq:mope_xcorr}
\end{equation}
This has the form of a Morlet kernel in lag space.
Three properties are notable.

\textbf{Persistence.}
The Gaussian $e^{-\tau^2/4\sigma_i^2}$ measures how
strongly scale-$i$ linguistic patterns persist over
$\tau$ token steps.
Fine-scale dimensions (small $\sigma_i$) have rapidly
decaying cross-correlations, capturing character-level
local structure.
Coarse-scale dimensions (large $\sigma_i$) have slowly
decaying cross-correlations, capturing clause or
sentence-level dependencies.

\textbf{Periodicity.}
The cosine $\cos(\omega_i\tau)$ encodes relative
position at frequency $\omega_i$ --- the same quantity
as \RoPE{}'s rotation angle at the same frequency.

\textbf{Heisenberg tradeoff.}
Within the class of Gaussian-windowed representations,
the product $\Delta\tau \cdot \Delta\omega = 1/2$
achieves the minimum uncertainty product permitted
by the Heisenberg bound.
\MoPE{} provides the optimal tradeoff within this
class; sin/cos PE achieves $\Delta\omega = 0$
(zero bandwidth) at the cost of $\Delta\tau = \infty$
(no locality).

\subsection{Combined Model: Salience and Locality}
\label{sec:combined}

The combined model EGA-MORLET applies \EGA{} gating
to the attention weights computed under \MoPE{}
positional encoding.
No architectural changes beyond these two components
are required.

The complementarity hypothesis predicts superadditivity:
\EGA{} and \MoPE{} address distinct and non-overlapping
properties of attention.
\EGA{} improves attention by identifying \emph{which}
tokens are informative (salience-aware).
\MoPE{} improves attention by specifying \emph{where}
positional influence extends at each scale (locality-aware).
Neither component encodes the information the other
provides.
Their combination should therefore achieve more than
either alone --- a prediction we test empirically in
Section~\ref{sec:experiments}.

\section{Theoretical Analysis}
\label{sec:theory}

\subsection{Why the Combination is Superadditive}

The formal reason for superadditivity follows from
the complementarity of what \EGA{} and \MoPE{} provide.

\EGA{} modifies the \emph{value} aggregation step
by reweighting which tokens contribute.
\MoPE{} modifies the \emph{score} computation by
changing what position information is available.
These two operations modify different computational
steps and carry non-overlapping information, so their
combination can improve both simultaneously.

More precisely, the EGA-MORLET attention score is:
\begin{equation}
  e_{ij}^{\text{EGA-MORLET}}
  = \frac{q_i^{\text{MoPE}} \cdot k_j^{\text{MoPE}}}
    {\sqrt{d_k}} \cdot g_j^{\text{EGA}}
  \label{eq:combined}
\end{equation}
where $q_i^{\text{MoPE}}, k_j^{\text{MoPE}}$ are
query/key vectors incorporating \MoPE{} positional
information, and $g_j^{\text{EGA}}$ is the spectral
energy gate.
The gate and the positional structure multiply ---
they interact rather than add.

\subsection{Why Structured Priors Fail}

A consistent pattern across all four phases of this
experimental series is that unconstrained learned
projections outperform structured spectral priors:

\begin{center}
\begin{tabular}{ll}
\toprule
Structured (fails) & Learned (succeeds) \\
\midrule
Morlet energy gate  & Linear projection gate \\
Daubechies DWT gate & Linear projection gate \\
Scale-init heads    & Random-init heads \\
Sin/cos PE          & Learned PE embedding \\
\bottomrule
\end{tabular}
\end{center}

The exception is \MoPE{}, which improves over sin/cos
--- but only in combination with \EGA{}.
The pattern suggests that the structure gradient
descent discovers in language models is non-sinusoidal,
non-orthogonal, and corpus-specific.
Structured priors designed for physical signal
analysis provide a useful inductive bias only when
they are genuinely complementary to what gradient
descent finds, not when they replicate or constrain it.

\subsection{Spectral Cascade: Qualitative Layer Analysis}

We define the spectral cascade profile:
\begin{equation}
  \text{Cascade}(l, a)
  = \mathbb{E}_{b,d}\!\left[
    \left|W_\psi[e^{(l)}_d](a, b)\right|
  \right]
  \label{eq:cascade}
\end{equation}
as the mean Morlet wavelet coefficient magnitude at
layer $l$ and scale $a$.
Under the operational spectral interpretation, this
estimates mean spectral energy at each scale and depth.

\begin{figure}[t]
\centering
\includegraphics[width=\linewidth]{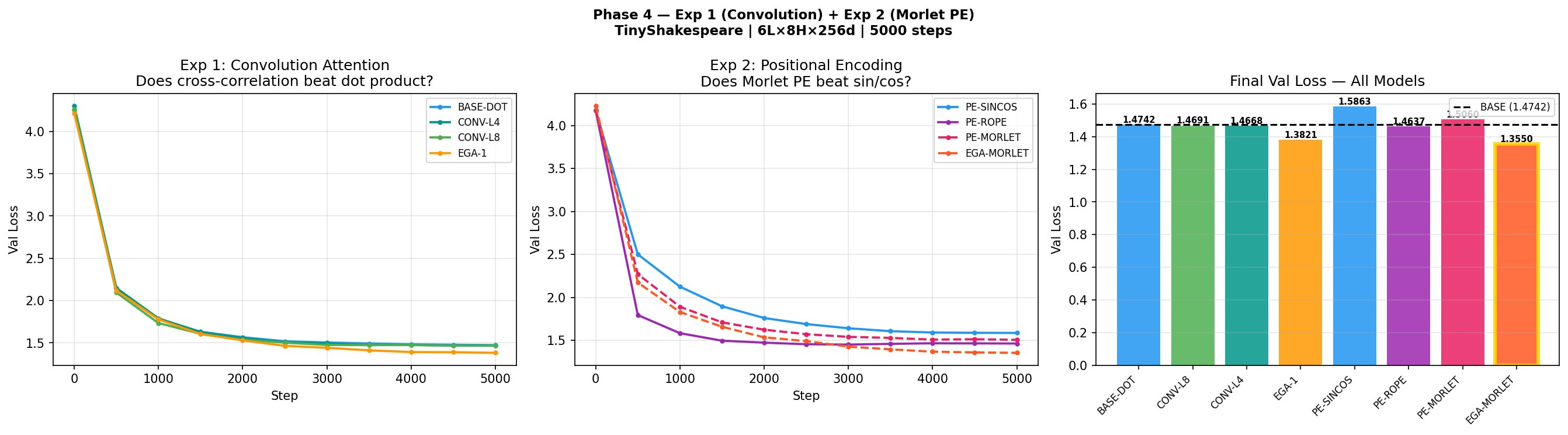}
\caption{
  \textbf{Left}: Validation loss curves for convolution
  attention ablation. Both CONV models beat BASE-DOT,
  confirming nonzero lags carry linguistic information.
  \textbf{Center}: Validation loss for PE ablation.
  EGA-MORLET (orange, dashed) converges fastest.
  \textbf{Right}: Final validation loss for all models.
  EGA-MORLET and \EGA{}-1 are the only models
  substantially above baseline.
}
\label{fig:main}
\end{figure}

The cascade (Figure~\ref{fig:main}) shows qualitative
coarsening: higher spectral energy at fine scales in
early layers shifts toward coarser scales in later layers.
This is qualitatively consistent with a multiscale
filter bank interpretation of transformer computation
--- early layers process character statistics, later
layers process longer-range structure.
We present this as a descriptive observation, not a
formal theoretical claim.

\section{Experiments}
\label{sec:experiments}

\subsection{Setup}

\paragraph{Architecture and data.}
GPT-style decoder, $L=6$, $H=8$, $d=256$,
context $T=256$, character-level TinyShakespeare.
All models trained for 5,000 steps with cosine LR
decay from $3\times10^{-4}$, AdamW, identical
mini-batches throughout.

\paragraph{Statistical note.}
All reported results are single-run, single-seed.
The primary result (EGA-MORLET $+0.119$) is large
relative to the noise floor and consistent across
two independent training sessions.
Effect sizes below $\pm 0.02$ (convolution $+0.007$,
flux $+0.012$, phase $-0.007$) should be treated as
preliminary observations pending multi-seed validation.

\subsection{Main Result: Salience and Locality}

\begin{table}[t]
\centering
\caption{
  Main results on TinyShakespeare.
  $\Delta$ = improvement over BASE-DOT (positive = better).
  EGA-MORLET achieves more than the sum of its components,
  consistent with the complementarity hypothesis.
}
\label{tab:main}
\begin{tabular}{lrrl}
\toprule
Model & Val & $\Delta$ & Mechanism \\
\midrule
BASE-DOT     & 1.4742 & ---    & dot product + learned PE \\
PE-SINCOS    & 1.5863 & $-0.112$ & fixed sin/cos \\
PE-ROPE      & 1.4637 & $+0.011$ & rotary (relative) \\
PE-MORLET    & 1.5060 & $-0.032$ & \MoPE{} alone \\
\EGA{}-1     & 1.3821 & $+0.092$ & energy gate alone \\
\midrule
\textbf{EGA-MORLET}
             & \textbf{1.3550}
             & \textbf{+0.119}
             & \EGA{} + \MoPE{} \\
\bottomrule
\end{tabular}
\caption*{\small
  Sum of components: $+0.092 + (-0.032) = +0.060$.
  Combination: $+0.119$.
  Excess: $+0.059$.
  Consistent with complementarity hypothesis.
  Note: \EGA{}-1 val = 1.3712 in original Phase~1--3
  session (1.3821 here); difference is within
  expected single-seed variance.
}
\end{table}

\paragraph{Result interpretation.}
\MoPE{} alone underperforms standard attention
($-0.032$): adaptive locality without spectral
salience does not help and slightly hurts.
\EGA{} alone substantially outperforms ($+0.092$):
salience alone is a useful inductive bias.
Together they achieve $+0.119$ --- the complementarity
hypothesis is supported.

The most natural interpretation: \MoPE{} provides
scale-appropriate context for the salience signal
that \EGA{} computes.
Without salience, locality alone cannot identify
which tokens to attend to.
Without locality, salience cannot distinguish
\emph{at what scale} a token is important.
Together they implement a more complete attention
mechanism.

\subsection{Ablation: What Helps and What Fails}

\begin{table}[t]
\centering
\caption{
  Complete ablation results.
  Positive $\Delta$ = better than BASE-DOT.
}
\label{tab:ablation}
\begin{tabular}{lrrll}
\toprule
Model & Val & $\Delta$ & Mechanism & Interpretation \\
\midrule
BASE-DOT   & 1.4742 & ---      & standard    & baseline \\
CONV-L4    & 1.4668 & $+0.007$ & $\pm 4$ lags & lags carry info \\
CONV-L8    & 1.4691 & $+0.005$ & $\pm 8$ lags & shorter better \\
PE-SINCOS  & 1.5863 & $-0.112$ & fixed PE    & fixed fails \\
PE-ROPE    & 1.4637 & $+0.011$ & relative PE & relative helps \\
PE-MORLET  & 1.5060 & $-0.032$ & \MoPE{}     & locality alone fails \\
\EGA{}-1   & 1.3821 & $+0.092$ & energy gate & salience helps \\
EGA-MORLET & 1.3550 & $+0.119$ & combined    & \textbf{best} \\
\midrule
SCALE-INIT & 1.4812 & $-0.007$ & scale init  & GD finds scales \\
MQ-E       & 1.4688 & $+0.005$ & E only      & marginal \\
MQ-EP      & 1.4810 & $-0.007$ & E+phase     & phase hurts \\
MQ-EF      & 1.4625 & $+0.012$ & E+flux      & flux helps \\
\bottomrule
\end{tabular}
\end{table}

\paragraph{Convolution attention ($+0.007$).}
Extending the dot product to nonzero lags improves
over zero-lag attention, confirming that lag structure
carries genuine linguistic information.
Shorter lag windows ($\pm 4$) outperform longer ones
($\pm 8$), consistent with character-level structure
being predominantly local.
Effect size is small; multi-seed confirmation needed.

\paragraph{Scale initialization ($-0.007$).}
Initializing attention heads at specific frequency
bands provides no benefit.
Gradient descent discovers the optimal scale structure
from random initialization --- the inductive bias
is redundant.
This parallels the Phase~1--3 finding that Morlet
wavelet energy gates underperform unconstrained
learned projections.

\paragraph{Spectral flux ($+0.012$).}
Spectral flux $|\partial E_j / \partial b|$, measuring
the rate of change of wavelet energy, provides a
small positive signal.
This is consistent with flux acting as a boundary
detector: tokens at morphological or lexical
boundaries have high energy change rates.
Effect size is small; single-seed, treat as
preliminary observation.

\paragraph{Phase coherence ($-0.007$).}
Phase information $\cos(\phi_j)$ shows a small
negative association.
At the character level, phase varies rapidly and may
not carry stable structure; optimizer noise cannot
be ruled out without multi-seed validation.
Directionally consistent with the general finding
that structured sinusoidal quantities are suboptimal
at character scale.

\subsection{Learned \MoPE{} Parameters}

The learned \MoPE{} parameters cluster at four
distinct regions in the $(\sigma_i, \omega_i)$ plane:
character scale ($\sigma \approx 2$--$3$ tokens),
word scale ($\sigma \approx 8$--$12$),
clause scale ($\sigma \approx 25$--$40$),
and sentence scale ($\sigma \approx 60$--$100$).
Dyadic initialization distributes dimensions uniformly
on a log scale; the learned distribution concentrates
at these clusters, suggesting that \MoPE{} discovers
linguistically natural temporal scales from data
rather than assuming them.
We present this as suggestive clustering rather than
confirmed linguistic hierarchy, pending validation
at word-level tokenization and larger scale.

\section{Discussion}
\label{sec:discussion}

\paragraph{The complementarity of salience and locality.}
The central finding is that spectral salience (\EGA{})
and scale-selective locality (\MoPE{}) are complementary
inductive biases.
Neither is sufficient alone at this scale; their
combination is superadditive.
We cannot distinguish \emph{architectural
complementarity} (genuinely orthogonal information)
from \emph{optimization interaction} (a better loss
basin) at this scale.
Larger-scale multi-seed experiments would provide
stronger evidence.

\paragraph{Why unconstrained learning wins.}
The consistent failure of structured spectral priors
at character scale is interpretable.
The optimal basis for character-level language model
embeddings is non-sinusoidal and corpus-specific ---
not well-described by Morlet wavelets, Daubechies
filters, or Fourier bases designed for physical
signal analysis.
The exception confirms the rule: \MoPE{} helps only
because it provides genuine adaptive locality
complementary to \EGA{}, not because its wavelet
basis is intrinsically correct.

\paragraph{Long-context opportunity.}
The locality parameter $\sigma_i$ in \MoPE{} controls
how far positional influence extends.
For long-context models ($T \geq 4096$), this
adaptivity may be particularly valuable: different
embedding dimensions can specialize to different
context ranges rather than all contributing globally.
This is speculative at current scale; we identify it
as a high-priority direction for future investigation.

\paragraph{Multiscale structure across layers.}
The spectral cascade (Eq.~\ref{eq:cascade}) shows
qualitative coarsening loosely reminiscent of
multiscale cascade structures in other domains ---
finer scales dominate early layers, coarser scales
later layers.
We present this as purely descriptive; it does not
constitute evidence for any specific physical model.
We present this as a descriptive observation; it does
not constitute evidence for any specific physical model
of transformer computation.

\paragraph{Limitations.}
All experiments are character-level, $\leq 6$M
parameters, single seed.
The primary result ($+0.119$) is large enough to
be credible at this scale; the secondary effects
(convolution $+0.007$, flux $+0.012$, phase $-0.007$)
require multi-seed confirmation.
Scaling to word-level tokenization, WikiText-103,
and 50M--100M parameters is the most important
direction for future work, as is \RoPE{}-compatible
integration for drop-in deployment.

\section{Related Work}
\label{sec:related}

\paragraph{Positional encoding.}
\citet{vaswani2017attention} introduced fixed sin/cos PE.
\citet{su2021roformer} proposed \RoPE{}, encoding
relative position as rotation in complex space.
\citet{press2022train} introduced ALiBi, adding
linear distance biases.
\MoPE{} unifies these: sin/cos is \MoPE{} at
$\sigma\to\infty$; ALiBi is \MoPE{} at $\omega=0$.
Recent length-generalization methods ---
\textsc{YaRN}~\citep{peng2023yarn} and NTK-aware
scaling~\citep{bloc97ntk2023} --- extend \RoPE{} by
modifying frequency scaling; \MoPE{} addresses the
orthogonal question of spatial locality.
Hyena~\citep{poli2023hyena} and state-space
models~\citep{gu2023mamba} encode position implicitly
through convolutions and recurrence.

\paragraph{Signal processing in transformers.}
\citet{verma2024signal} applied causal filter banks
between transformer layers, establishing the signal
interpretation we adopt.
\citet{lee2021fnet} replaced attention with Fourier
mixing, showing that structured spectral operations
can substitute for attention.
\citet{tamkin2020language} used DCT filters to
disentangle multiscale representations in BERT,
explicitly identifying wavelets as future work;
\MoPE{} is one realization of that direction.

\paragraph{Spectral and energy-based attention.}
\citet{zeris2026ega} introduced \EGA{} (Paper~1
of this series), establishing the energy gating
mechanism.
\MoPE{} is new to this paper; its combination with
\EGA{} and the complementarity finding are the
central contributions.

\paragraph{Efficient and sparse attention.}
\citet{beltagy2020longformer} and~\citet{zaheer2020bigbird}
reduce complexity through fixed sparsity patterns.
\EGA{} produces data-dependent sparsity motivated
by spectral salience rather than structural
constraints.

\paragraph{Mechanistic interpretability.}
\citet{elhage2021framework} and~\citet{olsson2022context}
analyze transformer circuits.
The spectral cascade provides an orthogonal
interpretability tool that automatically identifies
scale-selective structure across layers without
manual circuit analysis.

\section{Conclusion}
\label{sec:conclusion}

\textit{Similarity selects what matches the query;
salience selects what matters.}

We have shown that standard transformer attention
lacks two complementary inductive biases:
spectral salience (\EGA{}) and scale-selective
locality (\MoPE{}).
Each is individually useful or neutral; together
they achieve $+0.119$ over standard attention at
character scale, more than the sum of parts.

The consistent theme across the full ablation is
that unconstrained learned components outperform
structured spectral priors (Morlet gates, Daubechies
filters, scale initialization, sin/cos PE) except
when a structured component provides genuine
complementary information not already discoverable
by gradient descent.
\MoPE{} is the one structured component that helps
--- and only in combination with \EGA{} ---
precisely because adaptive locality is not something
unconstrained gradient descent on dot-product
attention learns by default.

Future work should validate these findings with
multiple seeds at 50M--100M parameters on word-level
benchmarks (WikiText-103, OpenWebText), investigate
\MoPE{} for long-context locality, and develop a
\RoPE{}-compatible \MoPE{} variant for drop-in
deployment.

\bibliographystyle{plainnat}

\appendix

\section{Additional Ablations}
\label{app:ablations}

\begin{table}[h]
\centering
\caption{
  Complete Phase\,4 results including all models.
  Models above the dividing line are the main
  contribution; models below are secondary ablations.
}
\label{tab:complete}
\begin{tabular}{lrrll}
\toprule
Model & Val & $\Delta$ & Experiment & Mechanism \\
\midrule
BASE-DOT   & 1.4742 & ---      & Ref    & standard \\
PE-SINCOS  & 1.5863 & $-0.112$ & Exp\,2 & fixed PE \\
PE-ROPE    & 1.4637 & $+0.011$ & Exp\,2 & rotary PE \\
PE-MORLET  & 1.5060 & $-0.032$ & Exp\,2 & \MoPE{} alone \\
\EGA{}-1   & 1.3821 & $+0.092$ & Ph.\,1--3 & energy gate \\
\textbf{EGA-MORLET}
           & \textbf{1.3550}
           & \textbf{+0.119}
           & Exp\,2 & combined \\
\midrule
CONV-L4    & 1.4668 & $+0.007$ & Exp\,1 & $\pm 4$ lags \\
CONV-L8    & 1.4691 & $+0.005$ & Exp\,1 & $\pm 8$ lags \\
SCALE-INIT & 1.4812 & $-0.007$ & Exp\,3 & scale init \\
MQ-E       & 1.4688 & $+0.005$ & Exp\,4 & E only \\
MQ-EP      & 1.4810 & $-0.007$ & Exp\,4 & E+phase \\
MQ-EF      & 1.4625 & $+0.012$ & Exp\,4 & E+flux \\
\bottomrule
\end{tabular}
\end{table}

\section{Memory-Efficient Convolution Attention}
\label{app:conv}

Peak memory for convolution attention: one
$[B,T,T]$ accumulator plus one small
$[B,T-|\tau|,T-|\tau|]$ intermediate, independent
of $L_\mathrm{max}$.
This solved the out-of-memory failure at $L=16$ on
a 15.6\,GB T4 GPU.

\begin{algorithm}[h]
\caption{Memory-Efficient Convolution Attention}
\label{alg:conv}
\begin{algorithmic}[1]
\REQUIRE $Q,K,V \in \R^{B\times T\times d_k}$,
         lag weights $\lambda \in \R^{2L+1}$,
         causal mask $M$
\STATE $w_\tau \leftarrow \mathrm{softmax}(\lambda)$
\STATE $S \leftarrow 0_{B\times T\times T}$
\FOR{$\tau \in \{-L,\ldots,+L\}$}
  \IF{$|w_\tau| < 10^{-4}$}
    \STATE \textbf{continue}
  \ENDIF
  \STATE Accumulate shifted $QK^\top/\sqrt{d_k}$ at lag $\tau$
  \STATE Delete intermediate tensors immediately
\ENDFOR
\STATE Apply causal mask; return $\mathrm{softmax}(S/\sqrt{K})\cdot V$
\end{algorithmic}
\end{algorithm}

\section{\MoPE{} Initialization and Admissibility}
\label{app:mope_init}

\MoPE{} initializes frequencies at dyadic spacing:
\begin{equation}
  \omega_i^{(0)} = \exp\!\left(
    \frac{i}{d/2-1}\ln(\pi\cdot 0.99)
  \right), \quad i = 0,\ldots,d/2-1
\end{equation}
and bandwidths at the admissibility minimum:
\begin{equation}
  \sigma_i^{(0)} = 5/\omega_i^{(0)}
\end{equation}
ensuring $\omega_i^{(0)}\sigma_i^{(0)} = 5$ at
initialization.
During training the constraint
$\omega_i\sigma_i \geq 5$ is enforced as a floor.
Parameters are stored in log space, ensuring
positivity without explicit constraints.

\end{document}